%% file: acl_latex.tex
\pdfoutput=1
\documentclass[11pt]{article}

\usepackage{authblk}
\usepackage[]{acl}

\usepackage{times}
\usepackage{latexsym}

\usepackage[T1]{fontenc}
\usepackage[utf8]{inputenc}

\usepackage{microtype}

\usepackage{subfigure}
\usepackage{graphicx}
\usepackage{amsmath}
\usepackage{multirow}
\usepackage{tabularx}
\usepackage{booktabs}
\usepackage{array}
\usepackage{cellspace}
\usepackage{amsfonts}
\usepackage{amssymb}
\usepackage{algpseudocode}
\usepackage{tabularray}
\usepackage[export]{adjustbox}
\usepackage{soul}
\usepackage{color}
\usepackage[ruled,linesnumbered]{algorithm2e}
\usepackage[T1]{fontenc}
\usepackage{mathtools}

\title{Unsupervised Scientific Abstract Segmentation with Normalized Mutual Information}

\author[$\dagger$]{\textbf{Yingqiang Gao}} 
\author[$\dagger$]{\textbf{Jessica Lam}}
\author[$\dagger \ddagger$]{\textbf{Nianlong Gu}} 
\author[$\dagger$]{\\ \textbf{Richard H.R. Hahnloser}}
\affil[$\dagger$]{Institute of Neuroinformatics, University of Zurich and ETH Zurich, Switzerland}
\affil[ ]{\{\texttt{yingqiang.gao, lamjessica, rich\}@ini.ethz.ch}}
\affil[$\ddagger$]{Linguistic Research Infrastructure, University of Zurich, Switzerland}
\affil[ ]{\{\texttt{nianlong.gu\}@uzh.ch}}

\begin{document}
\maketitle
\begin{abstract}
The abstracts of scientific papers consist of premises and conclusions.
Structured abstracts explicitly highlight the conclusion sentences, whereas non-structured abstracts may have conclusion sentences at uncertain positions.
This implicit nature of conclusion positions makes the automatic segmentation of scientific abstracts into premises and conclusions a challenging task.
In this work, we empirically explore using Normalized Mutual Information (NMI) for abstract segmentation.
We consider each abstract as a recurrent cycle of sentences and place segmentation boundaries by greedily optimizing the NMI score between premises and conclusions. 
On non-structured abstracts, our proposed unsupervised approach GreedyCAS achieves the best performance across all evaluation metrics; on structured abstracts, GreedyCAS outperforms all baseline methods measured by $P_k$. The strong correlation of NMI to our evaluation metrics reveals the effectiveness of NMI for abstract segmentation.\footnote{Code and data avaialble at \url{https://github.com/CharizardAcademy/abstract-segmentation.git}}
\end{abstract}

\section{Introduction}

In scientific papers, abstracts are short texts that summarize the findings written in the body text \cite{bahadoran2020principles}. A well-formulated abstract forms a clear pathway of scientific inference from the premises (e.g., shared knowledge, experimental evidence, or observation) to the conclusions (e.g., suggestions, claims) \cite{ripple2012structured}. Being able to split an abstract into a conclusion segment and a premise segment not only helps readers better comprehend how conclusions are drawn \cite{bahadoran2020principles} but also contributes to many downstream research tasks such as argument generation \cite{schiller2021aspect}, knowledge retrieval \cite{hua2019argument}, opinion analysis \cite{hulpus2019towards}, and text summarization \cite{cho2022toward}. 

\input{figure/cycle-format}

However, segmenting abstracts is not always easy. Many abstracts, especially those from the biomedical domain, are structured such that the reader can easily extract the conclusions (e.g., abstracts follow the IMRaD format \cite{nair2014organization, dernoncourt-lee-2017-pubmed}) or the CONSORT format \cite{hopewell2008consort}. 
In contrast, abstracts from many other research domains typically do not explicitly indicate the position of the conclusions, which means readers must perform the cognitively demanding task of identifying the conclusions themselves.
The difficulty in segmenting abstracts triggers the research interest in developing automatic approaches for splitting scientific abstracts into conclusion and premise sentences.

Existing text segmentation approaches \cite{ somasundaran2020two, lo2021transformer, barrow2020joint, koshorek2018text} can be applied to automatic scientific abstract segmentation. However, fine-tuning such models typically requires large amounts of labelled data that are expensive to collect. In comparison, unsupervised approaches require no annotated data and can provide massive amounts of segmented scientific abstracts with minimal human involvement. Thus, we focus on developing an unsupervised framework for scientific abstract segmentation.

To this end, we propose an explorative unsupervised framework for the task. Given a set of abstracts, we want to determine how each abstract should be split into a premise segment and a conclusion segment.
Combining the premises from all abstracts gives us a premise space, and similarly, we obtain a conclusion space. 
We hypothesize that the best segmentation for each abstract can be found when the \textbf{N}ormalized \textbf{M}utual \textbf{I}nformation (NMI) between the conclusion space and premise space is maximized. 
To approach the maximal NMI, we use an exhaustive greedy approach that iterates over all abstracts and determines the best segmentation for each abstract. 
To reduce the search space at each step, we first stitch the start and end of each abstract together to form a cycle, then select two segmentation boundaries with constraints based on our prior knowledge (see Figure~\ref{fig:cycle}). 
We name our approach the \textbf{Greedy} \textbf{C}ycled \textbf{A}bstract \textbf{S}egmentation (GreedyCAS) framework.

% such that the conclusions are most likely to be drawn from the premises. We quantitatively measure the likelihood of the conclusions using \textbf{N}ormalized \textbf{M}utual \textbf{I}nformation (NMI). 

% We achieve this goal by maximizing the reduced uncertainty of inferring conclusions from premises, which can be quantitatively measured by the \textbf{N}ormalized \textbf{M}utual \textbf{I}nformation (NMI) between them. We use NMI as the optimization objective to greedily find the segmentation of abstracts where the conclusions are most likely to be inferred from the premises. 
% We introduce the \textbf{Greedy} \textbf{C}ycled \textbf{A}bstract \textbf{S}egmentation (GreedyCAS) framework that operates on recurrent abstracts, in which we stitch the beginning of an abstract to its end to form a cycled structure. 

To test our proposed approach, we create two datasets. One dataset comprises non-structured abstracts with human-annotated conclusion sentences. The other dataset contains structured abstracts in which conclusion sentences are explicitly indicated.

% To test our proposed approach on both structured and non-structured abstracts, we create two datasets of scientific abstracts, one with human-annotated conclusion sentences, the other aggregated from structured abstracts in which conclusion sentences are explicitly highlighted.

Our main \textbf{contributions} are as follows:
\begin{itemize}
    \item We propose GreedyCAS, an unsupervised approach for scientific abstract segmentation that optimizes NMI.
    \item On the dataset of non-structured abstracts, we show that GreedyCAS achieves promising segmentation results.
    \item We find a strong correlation between NMI and our evaluation metrics, which proves the effectiveness of NMI being indicative of good segmentations.
\end{itemize}
% 1) ; 2) on the dataset of non-structured abstracts, we show that GreedyCAS achieves promising segmentation results; 3) we find a strong correlation between NMI and our evaluation metrics, which proves the effectiveness of NMI being reflective to good segmentations.

\section{Related Works}
Abstract segmentation is a particular case of text segmentation. The task of text segmentation is to insert separation markers into the text such that the segmented fragments are topically coherent and comprehensive \cite{hazem2020hierarchical}. Traditional methods can be categorized into supervised and unsupervised methods.

Unsupervised approaches \cite{alemi2015text} usually make use of metrics based on textual coherence or topic contiguity and take the following strategies: 1) greedily seeking the best segmentation at each step \cite{choi2000advances, hearst1994multi}; and 2) iteratively approaching the global optima of the segmentation objective via dynamic programming \cite{fragkou2004dynamic, bayomi2018c}. These methods usually directly utilize similarity measures or lexical frequencies between segments to determine segmentation boundaries or convert the inter-sentence similarities into a semantic graph and perform graph search \cite{glavavs2016unsupervised} to find the optimal segmentation. 

Supervised methods are usually deployed when sufficient annotated data is available. These methods typically use large language models to encode sentences and perform binary classification to predict whether each sentence falls on the segmentation boundary \cite{somasundaran2020two, banerjee2020segmenting, aumiller2021structural, badjatiya2018attention, lukasik2020text, koshorek2018text}. Despite having good performance, supervised approaches consume a large amount of fully annotated data which is expensive to collect.

\citet{banerjee2020segmenting} fine-tuned a hierarchical sentence encoder to classify sentences into discourse categories (\textit{Background}, \textit{Technique}, \textit{Observation}), which can be adapted to segment scientific abstracts.
However, their supervised approach requires structured abstracts as training data and thus was not designed for classifying sentences in non-structured abstracts. In this work, we aim to develop a generic approach that can be applied to both structured and non-structured abstracts.

 % that were used as premises for argument generation \cite{shieh2019towards}. 

% The authors \citet{banerjee2020segmenting} used transfer learning to train a hierarchical sentence encoder for the task. 

\input{figure/pipeline}

\section{Methodology}
\label{section:method}

We formulate the task of segmenting scientific abstracts as follows. Given an abstract $A = (s_i)_{i=1}^n$ containing $n$ sentences, we define $G^A = \{g_j^A\}_{j=1}^m$ as the space of all $m$ possible segmentations of $A$. 
Each segmentation $g_j^A = (P_j^A, C_j^A)$  contains a premise segment $P_j^A$ and a conclusion segment $C_j^A$ whose   segmentation boundaries are uniquely determined by two indices $\alpha^{A}_j, \xi^{A}_j \in \mathbb{Z}_{1:n}$ (range of integers from $1$ to $n$):
% that define the segmentation boundaries of the conclusion sentences:
\begin{align*}
    C_j^A &= \{ s_i \in A \mid \alpha^{A}_j \leq i \leq \xi^{A}_j \}, \\
    P_j^A &= \{ s_i \in A \mid s_i \notin C_j^A \}.
\end{align*}

For a corpus $\mathbb{A} = \{ A_i \}_{i \in \mathbb{Z}_{1:k}}$ of $k$ abstracts, we would have a set $\mathbb{G} = \{ G^{A_i} \}_{i \in \mathbb{Z}_{1:k}}$ of $m^k$ possible segmentations. 
Searching for the best segmentation for $\mathbb{A}$ within $\mathbb{G}$ as measured by an optimization objective involves exhaustively enumerating over $m^k$ segmentations, which is impossible under limited computational costs. Therefore, in this work, we concentrate on greedily approaching a reasonably good segmentation that is a tight lower bound of the actual global optima. 

\subsection{Cycled Abstract Configuration}

To reduce the search space for each abstract, we make use of our prior belief that conclusion sentences tend to be located at the start or the end of abstracts. This allows us to reshape each abstract as a cycle of sentences in which the end of the abstract meets its beginning and consider six (instead of $m$) possible segmentations per cycled abstract based on the following criteria: (1) each abstract contains at most three conclusion sentences; (2) conclusion sentences are located at the stitching point  (start or end position of the abstract) in the cycled abstract; (3) conclusion sentences are fenced by the segmentation boundaries $\alpha$ and $\xi$.

Table~\ref{tab:config-bound} demonstrates one example cycled abstract with seven sentences. For each sentence in the cycled abstract, if the sentence is the final sentence within the current segment, then we mark the sentence as a segmentation boundary. 

% Given the abstract corpus $\mathbb{A}$, our task is to find a set of segmenting configurations spanned by individual $G_{A_i}$, i.e. $\mathbb{G} = \{ G_{A_1}, \ldots, G_{A_k} \}$. The desired approximated optima $\mathcal{O}^*_{\mathbb{A}}$ for $k$ abstracts is reached, when for each $G_{A_i}$, an optimal $g_{A_i}^*$ is found one after the other, i.e.
% \begin{align*}
%     \mathcal{O}_{\mathbb{A}}^* = \argmax \mathcal{O}(\mathbb{G}^*) 
% \end{align*}
% where $\mathbb{G}^* = \{ g^*_{A_1}, \ldots, g^*_{A_k} \}$.

To fully utilize the power of parallel computing, we use multi-threading\footnote{We use the Python MultiThreading library \url{https://docs.python.org/3/library/threading.html}} for our greedy algorithm in our work to compute the optimization objective for different possible segmentations.
\input{table/config-boundary}

\subsection{Normalized Mutual Information}

Our next step is to choose an optimization objective for the greedy search. Inspired by the  previous work on text summarization \cite{padmakumar2021unsupervised} and birdsong analysis \cite{sainburg2019parallels}, we explore using mutual information as the optimization objective. As a probabilistic measure, mutual information $I(X; Y)$ indicates the absolute reduction of information uncertainty in bits for a random variable $X$ after observing the other correlated random variable $Y$; therefore, $I(X; Y)=0$ if and only if $X$ and $Y$ are independent. In our case, we aim to know how much uncertainty of observing the conclusions can be reduced with the presence of the premises. 

We denote $\mathbb{C} = \{ C_{j_i}^{A_i} \}_{i \in \mathbb{Z}_{1:k}}$ as one possible \textit{conclusion space} spanned by all conclusion segments from the $k$ abstracts, characterized by the segmentation boundaries $( j_i )_{i=1}^k$, and $\mathbb{P} = \{ P_{j_i}^{A_i} \}_{i \in \mathbb{Z}_{1:k}}$ being one possible \textit{premise space} obtained in the same way. Note that the segmentation number $j_i$ can be different for different abstracts $A_i$. 
Now the task can be formulated as follows: given the $k$ abstracts in $\mathbb{A}$, search for the particular premise space $\mathbb{P}$ and conclusion space $\mathbb{C}$ in which the mutual information $I(\mathbb{P}; \mathbb{C})$ is maximized. 

More formally, we compute $I(\mathbb{P}; \mathbb{C})$ as follows:
\begin{align*}
    & I(\mathbb{P}; \mathbb{C}) = \\% \sum_{P_{j_i}^{A_i} \in \mathbb{P}} \sum_{C_{j_i}^{A_i} \in \mathbb{C}} I(P_{j_i}^{A_i}; C_{j_i}^{A_i}) \\
    &  \sum_{A_i \in \mathbb{A}} \sum_{w_p \in P_{j_i}^{A_i}} \sum_{w_c \in C_{j_i}^{A_i}} p(w_p; w_c) \log \frac{p(w_p; w_c)}{p(w_p)p(w_c)}
\end{align*}
where $w_p$ and $w_c$ are unigram tokens in the $i$-th premise segment $P_{j_i}^{A_i}$ and the $i$-th conclusion segment $C_{j_i}^{A_i}$, respectively. $p(w_p; w_c)$ indicates the joint probability of the premise word $w_p$ appearing in the premise segment $P_{j_i}^{A_i}$ and the conclusion word $w_c$ appearing in the conclusion segment $C_{j_i}^{A_i}$, and $p(w_p)$ and $p(w_c)$ the marginal probability. Applying language modelling statistics, we compute the marginal probabilities as follows: 
\begin{align*}
    p(w_p) & = \frac{c(w_p, \mathbb{P})}{\sum_{w_p'} c(w_p', \mathbb{P})} \\ % \frac{|\{ w_p \in \mathbb{P}\}|}{\sum_{w \in \mathbb{P}} |\{ w \in \mathbb{P} \}| } \\
    p(w_c) & = \frac{c(w_c, \mathbb{C})}{\sum_{w_c'} c(w_c', \mathbb{C})} % \frac{|\{ w_p \in \mathbb{P}\}|}{\sum_{w \in \mathbb{P}} |\{ w \in \mathbb{P} \}| } \\
\end{align*}
where $c(w, \mathbb{P})$ denotes the number of occurrences of $w$ within the tokenized premise segments in $\mathbb{P}$ and $w_p'$ is a token from the premise segment of any abstract. The terms $c(w, \mathbb{C})$ and $w_c'$ are defined analogously.

The joint probability is then computed as
\[
    p(w_p; w_c) = 
    \frac{\sum_{i=1}^k c\bigl( w_p, P_{j_i}^{A_i} \bigr) c\bigl( w_c, C_{j_i}^{A_i} \bigr)}{\sum_{(w_p', w_c')} c(w_p', \mathbb{P}) c(w_c', \mathbb{C})},
\]

Mutual information is an unbounded measure that increases with the size of the abstract corpus $\mathbb{A}$ and is thus not comparable across different $\mathbb{P}$ and $\mathbb{C}$ \cite{poole2019variational}; therefore, we use a normalizing scalar to map $I(\mathbb{P}; \mathbb{C})$ onto the interval $[0,1]$ and use Normalized Mutual Information (NMI) as the final optimization objective.

Taken from \citet{kvaalseth2017normalized}, we compute NMI as follows:
\begin{equation*}
    \mathrm{NMI(\mathbb{P}; \mathbb{C})} = \frac{I(\mathbb{P}; \mathbb{C})}{\mathcal{U}_a}
\end{equation*}
where $\mathcal{U}_a$ denotes the non-decreasing theoretical upper bound of $I(\mathbb{P}; \mathbb{C})$ and is parametrized by the $a$-order arithmetic mean
\begin{equation*}
    \mathcal{U}_a= \left( \frac{\mathcal{U}_{\mathbb{P}}+ \mathcal{U}_{\mathbb{C}}}{2} \right)^{1/a}.
\end{equation*}
Here, we have
\begin{equation*}
    \mathcal{U}_{\mathbb{P}} = - \sum_{w_p} p(w_p) \log p(w_p) = H(\mathbb{P})
\end{equation*}
and
\begin{equation*}
    \mathcal{U}_{\mathbb{C}} = -\sum_{w_c} p(w_c) \log p(w_c) = H(\mathbb{C})
\end{equation*}
essentially being the entropy of the premise space and the conclusion space, respectively. For the least upper bound ($a=-\infty$), we have
\begin{equation*}
 \mathcal{U}_{-\infty} =  \lim_{a \rightarrow -\infty } \mathcal{U}_a = \min \{ \mathcal{U}_{\mathbb{P}}, \mathcal{U}_{\mathbb{C}} \}   
\end{equation*}
In this work, we use $\mathcal{U}_{-\infty}$ to normalize $I(\mathbb{P}; \mathbb{C})$ to ensure that the maximal attainable NMI value is 1. This brings us the benefit of having the NMI scores comparable for different corpus sizes $k$. 

\subsection{Greedy Cycled Abstract Segmentation}

We now introduce our GreedyCAS approach to segment abstracts of scientific papers. GreedyCAS performs an exhaustive search, where we first explore the best segmentation of one particular abstract that maximizes $\text{NMI}(\mathbb{P};\mathbb{C})$, then iterate over all abstracts for the same purpose. 

Algorithm~\ref{alg:greedy-base} describes the basic approach \textit{GreedyCAS-base}. Given the input abstracts $\mathbb{A}$, the algorithm greedily searches for the segmentation that leads to the maximal $\text{NMI}(\mathbb{P};\mathbb{C})$. The output is the optimized segmentation $\mathbb{G}^*$ that contains the best segmentation for individual abstracts. 

\input{table/greedy-base}

Algorithm~\ref{alg:greedy-plus} illustrates the advanced approach \textit{GreedyCAS-NN}, where we first split the abstract corpus $\mathbb{A}$ into a series of chunks; then, for each seed abstract $A^s_{j_i}$ sampled from the current chunk, we perform embedding-based nearest neighbour (NN) search within the chunk to construct the batch comprising the most semantically relevant abstracts to $A^s_{j_i}$. The most semantically relevant abstracts of $A^s_{j_i}$ are fetched according to the cosine similarities calculated using their abstract embeddings. Finally, the same greedy process is performed to find the best segmentation for each abstract.

\input{table/greedy-plus}

\section{Dataset}

Since we calculate NMI scores using the lexical co-occurrences of the words, to have reliable estimations of the probabilities, and to obtain well-parsed scientific abstracts, we use the COVID-19 Open Research Dataset (CORD-19) released by \citet{wang-etal-2020-cord}. This dataset is a massive collection of scientific papers on SARS-CoV-2 coronavirus research published since March 2020. These papers share higher lexical commonality than biomedical papers in general due to the focused research interests in COVID-19.

To collect structured abstracts, we then work with abstracts that are structured into \textit{Background}, \textit{Methods}, \textit{Results}, and \textit{Conclusion} categories. We trust the categorization of these structured abstracts from the CORD-19 corpus since scientific papers are peer-reviewed and multi-round revised. We automatically aggregate the dataset \textbf{CAS-auto} from 697 structured scientific abstracts whose paper title contains the keyword ``vaccine''. Inspired by \citet{shieh2019towards}, we take sentences in \textit{Background}, \textit{Methods}, \textit{Results} categories as premises, and sentences in \textit{Conclusion} category as conclusions.

In addition, we manually construct a dataset \textbf{CAS-human} of 196 non-structured abstracts from CORD-19, using the keyword ``antigen'' to filter out target abstracts. We then ask four human annotators to label the conclusion sentences within those abstracts. All human annotators were not instructed about the potential positions of conclusion sentences in scientific abstracts. By doing this, we minimize the bias introduced to the human annotators. To facilitate the annotation process and reduce the annotators' workload, we use the interactive data labelling platform Doccano\footnote{MIT License, available at~\url{https://github.com/doccano/doccano}} \cite{doccano} for constructing the CAS-human dataset.

Table~\ref{tab:stats} shows the overall statistics of our proposed datasets for scientific abstract segmentation. During the data preprocessing, we intentionally remove stop words, numbers, and punctuations (except ``.'', which is essential for the sentence tokenizer\footnote{We use the sentence-splitter by Philipp Koehn and Josh Schroeder, GNU Lesser General Public License, available at \url{https://github.com/mediacloud/sentence-splitter}} we use) in the abstracts for meaningful and reliable word counts. We also lowercase all tokens in both datasets for better computational efficiency. 

\input{table/stats}

Figure \ref{fig:con-pos-stat} shows the positions of the conclusion sentences in the CAS-human dataset as labelled by the human annotators. Similarly as shown in previous works \cite{fergadis2021argumentation, achakulvisut2019claim}, 95\% of our annotated non-structured abstracts, the positions of conclusion sentences are consistent with our prior knowledge. 

\input{figure/con-pos-stat}

\section{Evaluation}

To evaluate the segmentation results, we propose using metrics for text segmentation, set similarity and semantic relevance. 

\subsection{Metrics}
We test the performance of our approaches on both automatically (CAS-auto) and manually (CAS-human) aggregated data. We use text segmentation metrics $P_k$ \cite{beeferman1999statistical} and WindowDiff (WD, \citet{pevzner2002critique}) to evaluate the segmentation boundaries for each abstract. Then, we use ROUGE score \cite{lin-2004-rouge} to measure the semantic relevance between the segmented and ground-truth conclusion sentences. To quantify the textual relevance, we compute the arithmetic mean of ROUGE-1, ROUGE-2, and ROUGE-Lsum f-measures. Finally, we use Jaccard index to measure the similarity between the set of segmented conclusion sentences and the set of ground truth conclusion sentences. Lower $P_k$ and WD score indicate better segmentation results, whereas higher ROUGE and Jaccard index represents better segmentation results.

\subsection{Baselines}
%We introduce three unsupervised baseline approaches for cycled abstract segmentation as baselines. To make the results comparable, for each of the baseline approaches, we manually place one additional segmentation boundary at the end of the abstract, if the approach provides only one segmentation boundary.

We present three unsupervised baseline methods for cycled abstract segmentation as baselines. To ensure comparability of the results, we manually add an additional segmentation boundary at the end of the abstract for any approach that provides only one boundary.

\paragraph{Random} To test our prior knowledge of the position of conclusions sentences, we set up two random baselines: a) \textit{Random-base} following the idea initially proposed by \citet{beeferman1999statistical}, we place segmentation boundaries after two randomly selected sentences; b) \textit{Random-plus}, we segment a given abstract by randomly selecting one from the six possible segmentations. 

\paragraph{TextTiling\footnote{We use the \textit{HarvestText} implementation for English language, MIT License, available at \url{https://github.com/blmoistawinde/HarvestText}}} As proposed in \citet{hearst1997text} and served as the classic text segmentation approach, TextTiling utilizes lexical information to detect topic changes within a given text. In our case, TextTiling places a segmentation boundary between sentences.

\paragraph{SBERT-sim\footnote{We use the \textit{sentence-transformers} for Sentence-BERT encoder (pretrained model \textit{all-MiniLM-L6-v2} with model size 80 MB), Apache-2.0 License, available at \url{github.com/UKPLab/sentence-transformers}}} As inspired by \cite{solbiati2021unsupervised}, we use a pre-trained Sentence-BERT \cite{reimers-2019-sentence-bert} text encoder to segment abstracts using sentence semantics. Each abstract is split into two segments such that the cosine similarity of their Sentence-BERT embeddings is maximized. 

\section{Results and Discussion}

\subsection{Non-structured Abstracts}

First, we test our supervised GreedyCAS-approaches against the baselines on the CAS-human dataset. We list the experiment results in Table \ref{tab:results-human}. We use the same pre-trained Sentence-BERT encoder as the SBERT-sim baseline for GreedyCAS-NN. For the GreedyCAS approaches, we report the empirical performance with the best batch size. The best batch size hints at how many closely related abstracts together can achieve the highest NMI score. Intuitively as the batch size grows, the relatedness among the abstracts drops because it is not possible to get a large number of abstracts that study precisely the same research question.

\input{table/results-human}

We see that the GreedyCAS-NN achieves the best performance against all other approaches. The results suggest that our approaches work well on non-structured abstracts. Thus, formulating abstracts into cycled structures helps capture the conclusion-relevant information at both the abstracts' beginning and end.

\subsection{Structured Abstracts}

Next, we test GreedyCAS approaches against the baselines on the CAS-auto dataset. The results are shown in Table \ref{tab:results-auto}. 

\input{table/results-auto}

Firstly, we found that compared to the Random-base baseline,  Random-plus improves the results across different measures by large margins. This verifies our prior belief that treating an abstract's last three and the first two sentences as conclusion sentences helps reduce the time complexity while maintaining good performance. 

We then observe that the SBERT-sim baseline, which segments abstracts based on the cosine similarity between the premise and conclusion segments, achieves the best performance on three out of four metrics by a large margin. Our best model GreedyCAS-NN only achieves the leading performance measured by $P_k$, while achieving lower performance measured by other metrics. 
In Table~\ref{tab:structured-example-A} and~\ref{tab:structured-example-B} in appendix~\ref{sec:appendix-example}, we show two abstracts that were wrongly segmented by GreedyCAS-NN: the first abstract has one additional sentence from the \textit{Background} category, whereas the second abstract has one additional sentence from the \textit{Results} category.
These sentences were misattributed by GreedyCAS to the conclusion segment because they increase the NMI score; however, due to the complexity of NMI, understanding the exact reasons why NMI increases is non-trivial. In appendix~\ref{sec:appendix-hardness}, rather than study the complex scenario where an entire sentence gets re-attributed, we showcase the difficulty by considering the impact on NMI when just one word is moved from the premise segment to the conclusion segment.

\subsection{Analysis}

We investigate whether NMI is indicative of better segmentation results carried out by GreedyCAS-NN. We test this on the CAS-human dataset. We compute the correlation coefficients between NMI scores and each evaluation metric. We plot the NMI scores and the evaluation metrics w.r.t the batch size (ranging from 2 to 12).

\input{figure/correlation}

Figure \ref{fig:correlation} shows that NMI scores are strongly negatively correlated with the text segmentation metrics $P_k$ and WindowDiff, but are strongly positively correlated with the set similarity metric Jaccard index and the semantic metric ROUGE. 

We also studied whether placing segmentation boundaries between words can provide similar $\text{NMI}(\mathbb{P};\mathbb{C})$ scores compared to placing them within sentences. To do this, we randomly pick one abstract from the CAS-human dataset and calculate the changes in $\text{NMI}(\mathbb{P};\mathbb{C})$ due to moving the segmentation boundary by one word to the right, where we segment the remaining abstracts with the ground truth. 

\input{figure/delta-mi}

We see in Figure~\ref{fig:delta-mi} that for each sentence, the maximal $\text{NMI}(\mathbb{P};\mathbb{C})$ is achieved near or at the sentence boundary. 
This suggests that it is not necessary to segment abstracts on the word level, which takes much longer than segmenting on the sentence level.

\section{Conclusion}

In this explorative work, we propose an unsupervised approach, GreedyCAS, that automatically segments scientific abstracts into conclusions and premises. We introduce the cycled abstract segmentation pipeline, which can be applied to structured and non-structured abstracts. Our approach leverages the lexical information between words that co-occur in the conclusion and premise segments and finds the best segmentation of a set of abstracts using NMI as an optimization objective. Our empirical results show that NMI is an effective indicator for the segmentation results of scientific abstracts.   

\section{Limitations}
In this work, we explored the use of normalized mutual information as an optimization objective for scientific abstract segmentation. The main limitations of our work are listed below: 
\begin{itemize}
\item The input abstracts of GreedyCAS need to be on similar research topics, otherwise, their shared vocabulary is limited and the word probabilities in the computation of NMI cannot be estimated well. 
We have also tested GreedyCAS on the abstracts of biomedical papers from the S2ORC dataset \cite{lo2020s2orc}, which have diverse research topics and are not restricted to COVID-19, but obtained less promising results. 
\item GreedyCAS has a high time and space complexity because it involves an exhaustive search for the best segmentation and enumeration over all possible word pairs at each iteration. As a result, GreedyCAS takes a long time to execute and using larger batch sizes is a challenge with limited computational resources. 
We attempted to implement a faster and leaner approximated version of GreedyCAS by using only the most influential word pairs but were unable to do so.
\end{itemize}

In future, we will analytically study how to increase the efficiency and the applicability of GreedyCAS by considering other ways of estimating the word probabilities. 

%\section{Acknowledgements}

%We acknowledge the support from Swiss National Science Foundation NCCR Evolving Language, Agreement No.51NF40\_180888. 
%We also thank the anonymous reviewers for their constructive comments and feedback.

\bibliography{anthology,custom}
\bibliographystyle{acl_natbib}

\newpage
\appendix
\onecolumn

\section{Dataset Example}
\label{sec:appendix-example}

In the following tables, sentences in the premise segment are highlighted in blue, whereas sentences in the conclusion segment are highlighted in red. 

\input{table/structured-example-A}
\input{table/structured-example-B}
\input{table/word-pair-example}

\section{Impact on NMI when Moving One Word}
\label{sec:appendix-hardness}

Let $w_p \in \mathbb{P}$ be any premise word, and $w_c \in \mathbb{C}$ be any conclusion word. Given a fixed abstract $A_i$ and one possible segmentation $g_j^{A_i}$ (we use the index $j$ to represent the segmentation). By moving one arbitrary word $w$ in abstract $A_i$ from the premise segment $P^{A_i}_j$ to the conclusion segment $C^{A_i}_j$, we try to infer how $I(\mathbb{P}; \mathbb{C})$ changes. We investigate the major terms in the equation of mutual information. We aim at finding those word pairs that predominantly contribute to $I(\mathbb{P};\mathbb{C})$ so that the computation can be simplified, which essentially will allow the algorithm to run with a larger batch size. 

First, we examine what happens to the marginal probability $p(w_p)$ and $p(w_c)$:
\begin{align*}
    p(w_p) &= \frac{c(w_p, \mathbb{P}) - \mathbb{I}[w_p = w]}{\sum_{w_p'} c(w_p', \mathbb{P}) - 1} \\
    p(w_c) &= \frac{c(w_c, \mathbb{C}) + \mathbb{I}[w_c = w]}{\sum_{w_c'} c(w_c', \mathbb{C}) + 1}
\end{align*}

Here $\mathbb{I}$ denotes an indicator function and $c$ a counter function. The indicator function takes the value 1 if the condition in the bracket is fulfilled, otherwise, it takes the value 0. For the marginal probabilities, we have the following cases:
\begin{itemize}
    \item If $w_p = w$, then $p(w_p)$ decreases; if $w_p \neq w$, then $p(w_p)$ increases.
    \item If $w_c = w$, then $p(w_c)$ increases; if $w_c \neq w$, then $p(w_c)$ decreases.
\end{itemize}

Then, we examine what happens to $p(w_p; w_c)$, the main term in computation of $I(\mathbb{P}; \mathbb{C})$
\begin{align*}
    &p(w_p; w_c) \\
    &= \frac{\sum_{j \neq i} c(w_p, P^{A_j}) c(w_c, C^{A_j}) + \left( c(w_p, P^{A_i}) - \mathbb{I}[w_p = w] \right) \left( c(w_c, C^{A_i}) + \mathbb{I}[w_c = w]\right)}{\sum_{(w_p', w_c')} \left( c(w_p', \mathbb{P}) - \mathbb{I}[w_p' = w] \right) \left( c(w_c', \mathbb{C}) + \mathbb{I}[w_c' = w] \right)} \\
    &= \frac{\overbrace{\sum_{j \neq i} c(w_p, P^{A_j}) c(w_c, C^{A_j})}^{\text{constant}} + \overbrace{\left( c(w_p, P^{A_i}) - \mathbb{I}[w_p = w] \right)}^{\alpha} \overbrace{\left( c(w_c, C^{A_i}) + \mathbb{I}[w_c = w]\right)}^{\beta}}{\underbrace{\sum_{(w_p', w_c')} c(w_p', \mathbb{P}) c(w_c', \mathbb{C})}_{\text{constant}} + \underbrace{\sum_{(w_p', w_c')} \left\{ c(w_p', \mathbb{P}) \mathbb{I}[w_c' = w] - c(w_c', \mathbb{C}) \mathbb{I}[w_p' = w] - \mathbb{I}[w_p' = w] \mathbb{I}[w_c' = w] \right\}}_{\gamma}} \\
    &= \frac{a + \alpha \beta}{b + \gamma},
\end{align*}

here $a$ and $b$ are the constant terms within the fraction, since moving $w$ in $A_i$ will not affect other abstracts. For the joint probability, we have the following cases:
\begin{itemize}
    \item If $w_p \neq w$ and $w_c \neq w$, $p(w_p;w_c)$ remains unchanged.
    \item If $w_p = w$ and $w_c \neq w$, then

$$
    p(w_p; w_c) 
    = \frac{a + (\alpha-1)\beta}{b - c(w_c, \mathbb{C})}.
$$
    \item If $w_p \neq w$ and $w_c = w$, then

$$
    p(w_p; w_c) 
    = \frac{a + \alpha(\beta+1)}{b + c(w_p, \mathbb{P})}.
$$
    \item If $w_p = w$ and $w_c = w$, then

$$
    p(w_p; w_c) 
    = \frac{a + (\alpha-1)(\beta+1)}{b + c(w_p, \mathbb{P}) - c(w_c, \mathbb{C}) - 1}.
$$
\end{itemize}

Till this point, we found it very difficult to predict how $p(w_p;w_c)$ will change when moving $w$. The reasons are: 

\begin{itemize}
    \item For the case of $w_p = w$ and $w_c = w$, we cannot tell a priori whether $c(w, \mathbb{P}) - c(w, \mathbb{C})$ is positive, i.e. whether $w$ appears more frequently within premise segments or conclusion segments. This leads to the uncertainty of determining the change in the sign of mutual information.
    \item The normalizing factor of NMI, which is essentially the minimum between $H(\mathbb{P})$ and $H(\mathbb{C})$, cannot be determined after moving $w$. 
\end{itemize}

Also, because for any research domain, it is nearly impossible to get a large number of papers that study exactly the same research question (e.g., it's not possible to get thousands of papers that study the effectiveness of COVID-19 vaccines), therefore, further increasing the batch size is not feasible.

Due to the above reasons, we only computationally studied how NMI would change when moving one word from the premise segment to the conclusion segment (see Figure~\ref{fig:delta-mi}).

\end{document}

%% file: figure/cycle-format.tex
\begin{figure}[!htb]
\centering
\includegraphics[width=\columnwidth]{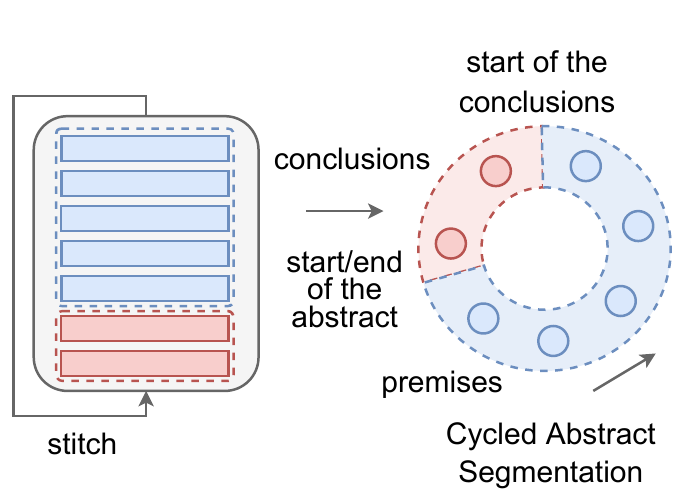}
\caption{\textbf{Left}: An abstract that contains premise (blue) and conclusion (red) sentences. The task is to identify the start and end segmentation boundaries of the conclusions. \textbf{Right}: We regard each abstract as a recurrent cycle of sentences by stitching its start and end together. Best view in color printing.}
\label{fig:cycle}
\end{figure}

%% file: figure/pipeline.tex
\begin{figure*}[!htb]
\includegraphics[width=\textwidth]{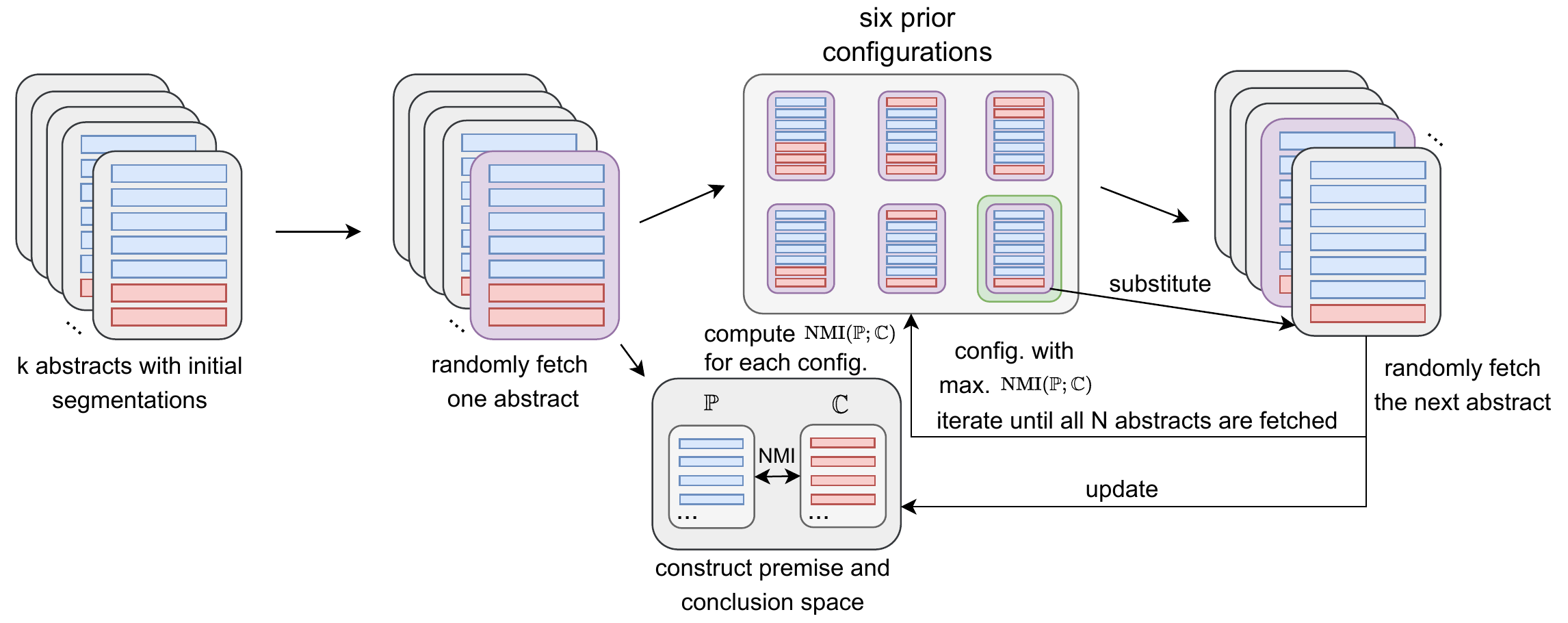}
\caption{Our proposed GreedyCAS pipeline for segmenting scientific abstracts. Premise sentences are colored in blue, whereas conclusion sentences are colored in red. Best view in color printing.}
\label{pipeline}
\end{figure*}

%% file: table/config-boundary.tex
\newcommand\cincludegraphics[2][]{\raisebox{-0.4\height}{\includegraphics[#1]{#2}}}

\begin{table}[!thb]
    \centering
    \resizebox{\columnwidth}{!}{
    \begin{tabular}{p{1.7cm}ccc}
    \toprule
    \multirow{1}{*}{config.} &
    \multicolumn{1}{c}{\cincludegraphics[width=0.1\textwidth]{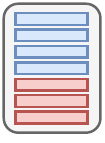}}
     & \multicolumn{1}{c}{\cincludegraphics[width=0.1\textwidth]{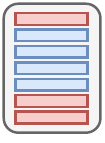}}  & \multicolumn{1}{c}{\cincludegraphics[width=0.1\textwidth]{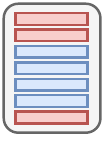}} \\
     \midrule
    labelling & 0001001 & 1000100 & 0100010 \\ 
    \midrule
    \multirow{1}{*}{config.} & \multicolumn{1}{c}{\cincludegraphics[width=0.1\textwidth]{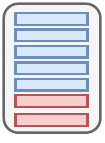}} & \multicolumn{1}{c}{\cincludegraphics[width=0.1\textwidth]{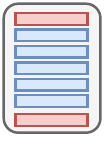}} & \multicolumn{1}{c}{\cincludegraphics[width=0.1\textwidth]{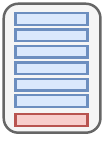}} \\
    \midrule
    labelling & 0000101 & 1000010 & 0000011 \\
    \bottomrule
    \end{tabular}
    }
    \caption{Labelling of segmentation boundaries for an example abstract with seven sentences. 1 indicates a sentence on the segment boundary, whereas 0 indicates a non-boundary sentence.}
    \label{tab:config-bound}
\end{table}

%% file: table/greedy-base.tex
\let\emptyset\varnothing

\begin{algorithm}[!htb]
\caption{GreedyCAS-base: unsupervised cycled abstract segmentation}\label{algorithm}
\label{alg:greedy-base}

\small
\KwIn{abstract corpus $\mathbb{A} = \{ A_i \}_{i \in \mathbb{Z}_{1:k}}$}
\KwOut{optimized segmentation $\mathbb{G} = \{ G^{A_i} \}_{i \in \mathbb{Z}_{1:k}}$}

$\mathbb{A}^{\text{res}} \leftarrow \mathbb{A}$\;
\While{$\mathbb{A}^{\text{res}} \neq \varnothing$}{
    $\mathcal{O}_{\mathbb{A}}^* \leftarrow 0$; \Comment{optimization objective}.\\
    $\mathbb{G}^* \leftarrow \varnothing$;
    
    $A_i \leftarrow \text{sample}(\mathbb{A})$\;
        
    $\mathbb{A}^{\text{res}} \leftarrow \mathbb{A}^{\text{res}} \backslash A_i$\;
    
    $\{ g_{j}^{A_i} \}_{j=1}^6 \leftarrow \text{configure}(A_i)$\;
    
    $G^{A_i} \leftarrow \{ g_{j}^{A_i} \}_{j=1}^6$\;
    
    \ForEach{epoch}{
        $\mathbb{G}^{\mathbb{A}^{\text{res}}} \leftarrow \varnothing$\;
        
        \ForEach{$A_r \in \mathbb{A}^{\text{res}} $}{
           $ \{g_j^{A_r} \}_{j=1}^6 \leftarrow \text{configure}(A_r)$\;
           
           $g^{A_r}_{j} \leftarrow \text{sample}(\{g_j^{A_r}\}_{j=1}^6)$\;
           
           $G^{A_r} \leftarrow \{ g_j^{A_r} \}$\;
           $\mathbb{G}^{\mathbb{A}^{\text{res}}} \leftarrow \mathbb{G}^{\mathbb{A}^{\text{res}}} \cup G^{A_r} $\;
        }
        
        \ForEach{$g^{A_i}_{j} \in G^{A_i}$}{
            $g^{A_i}_{j} = (P^{A_i}_{j}, C^{A_i}_{j})$\;
            
            \ForEach{$G^{A_r} \in \mathbb{G}^{\mathbb{A}^{\text{res}}}$}{
                $ g_{j}^{A_r} \leftarrow G^{A_r}$\;
                $g^{A_r}_j =( P^{A_r}_j, C^{A_r}_j )$\;
            }

            $\mathbb{P} = P^{A_i}_j \cup \{ P_{j}^{A_r} \}_{A_r \in \mathbb{A}^\text{res}}$\;

            $\mathbb{C} = C^{A_i}_j \cup \{ C_{j}^{A_r} \}_{A_r \in \mathbb{A}^\text{res}}$\;
            
            $\mathcal{O}_{\mathbb{A}} \leftarrow \text{compute-NMI}(\mathbb{P}; \mathbb{C})$\;
            
            \If{$\mathcal{O}_{\mathbb{A}} > \mathcal{O}_{\mathbb{A}}^*$}{
                $\mathcal{O}_{\mathbb{A}}^* \leftarrow \mathcal{O}_{\mathbb{A}}$\;
                
                $\mathbb{G}^* \leftarrow \mathbb{G}^{\mathbb{A}^{\text{res}}} \cup G^{A_i}$\;
                
            }
            
            \Else{
            continue; 
            }
        }
    }
}

\Return{$\mathbb{G}^*$\;}
\end{algorithm}

%% file: table/greedy-plus.tex
\let\emptyset\varnothing

\begin{algorithm}[!htb]
\caption{GreedyCAS-NN: unsupervised cycled abstract segmentation with nearest neighbor search}
\label{alg:greedy-plus}
\small
\KwIn{abstract corpus $\mathbb{A} = \{ A_i \}_{i \in \mathbb{Z}_{1:k}}$, chunk size $c$, batch size $b$}
\KwOut{optimized segmentation $\mathbb{G}^* = \{ G^{A_i} \}_{i \in \mathbb{Z}_{1:k}}$}

$\mathbb{G}^* \leftarrow \varnothing$\;

$\{ \mathbb{A}^{\text{chunk}}_i \}_{i \in \mathbb{Z}_{1:k/c}} \leftarrow \text{truncate}(\mathbb{A}, c)$\;

$\mathbb{A}^{\text{chunk}} \leftarrow \{ \mathbb{A}^{\text{chunk}}_i \}_{i \in \mathbb{Z}_{1:k/c}}$\;

\While{$\text{len}(\mathbb{G}^*) \neq k$}{
    \ForEach{$\mathbb{A}^{\text{chunk}}_i \in \mathbb{A}^{\text{chunk}}$}{
    $\mathbb{G}^{\text{chunk}}_* \leftarrow \varnothing$\;
    
    $\{ A^s_{j_i} \}_{j \in \mathbb{Z}_{1:c/b}} \leftarrow \text{sample}(\mathbb{A}^{\text{chunk}}_i)$\;
    
    $ \mathbb{A}^s_i \leftarrow \{ A^s_{j_i} \}_{j \in \mathbb{Z}_{1:c/b}}$; \Comment{seed abstracts}
    
        \ForEach{$A^s_{j_i} \in \mathbb{A}^s_i$}{
            $\{ A_{j_im}^s \}_{m \in \mathbb{Z}_{1:b}} \leftarrow \text{NN-search}(A^s_{j_i};b)$\; 
            $\mathbb{A}^{\text{batch}}_s \leftarrow \{ A_{j_im}^s \}_{m \in \mathbb{Z}_{1:b}}$\;
            
            $ \mathbb{G}^{\text{batch}}_* \leftarrow \text{GreedyCAS-base}(\mathbb{A}^{\text{batch}}_s)$\;
            $\mathbb{G}^{\text{chunk}}_* \leftarrow \mathbb{G}^{\text{chunk}}_* \cup \mathbb{G}^{\text{batch}}_*$\;
            }
        $\mathbb{G}^* \leftarrow \mathbb{G}^* \cup \mathbb{G}^{\text{chunk}}_*$\;
    }
    
}
\Return{$\mathbb{G}^*$\;}

\end{algorithm}

%% file: table/stats.tex
\begin{table}[!htb]
\resizebox{\columnwidth}{!}{
\begin{tabular}{lcccc}
\toprule
Dataset & \#abs. & \#con. & \#pre. & avg. $|\text{abs.}|$ \\
\midrule
CAS-auto  & 697 & 1,267 & 4,755 & 8.64 \\
CAS-human & 196 & 263 & 1,220 & 7.57 \\
\bottomrule
\end{tabular}
}
\caption{Statistics of the two datasets. \#abs denotes the number of abstracts, \#con. and \#pre. indicate the number of conclusion and premise sentences, respectively, and avg. $|\text{abs.}|$ the average number of sentences.}
\label{tab:stats}
\end{table}

%% file: figure/con-pos-stat.tex
\begin{figure}[!htb]
    \centering
    \includegraphics[width=0.9
    \columnwidth]{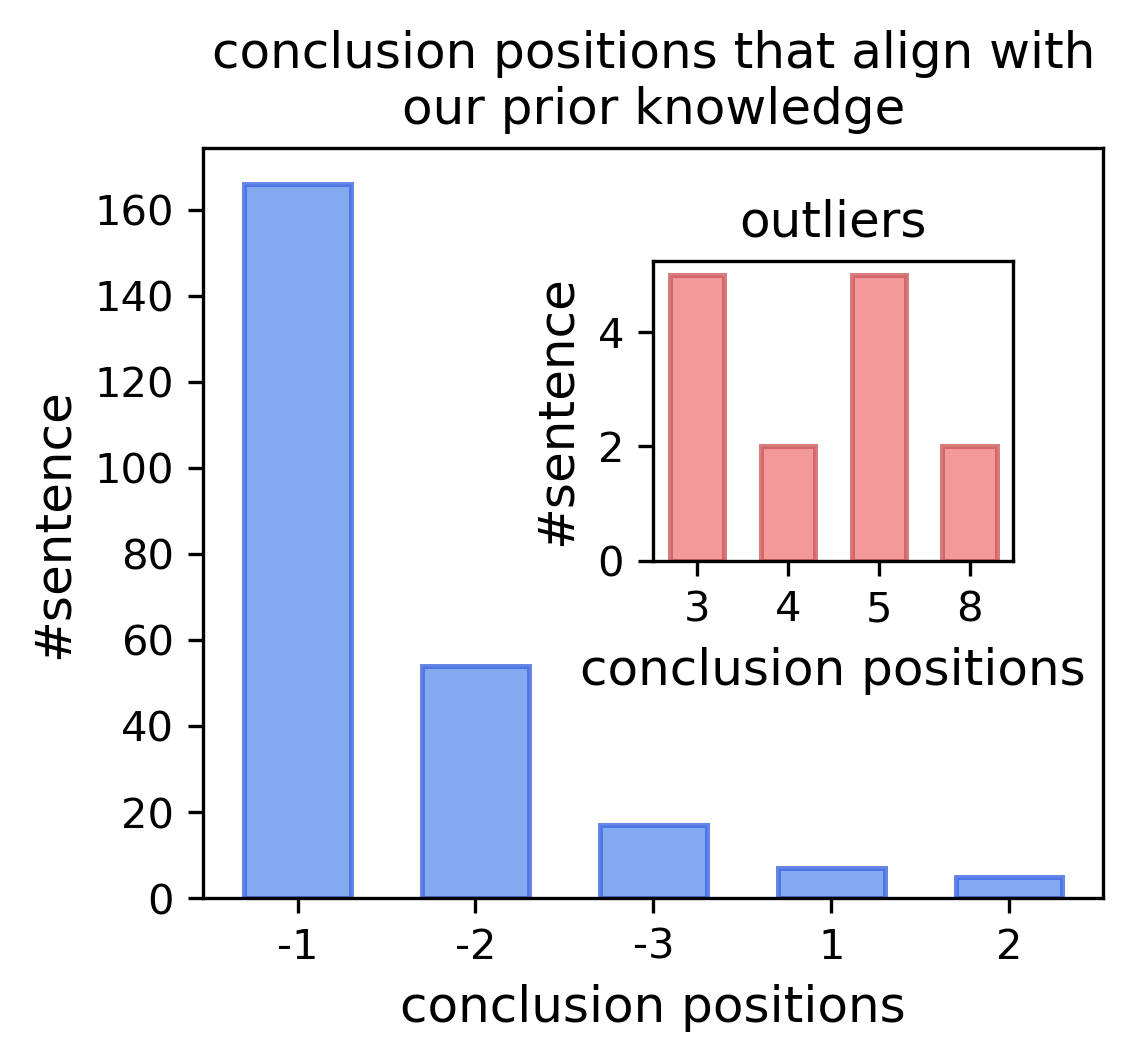}
    \caption{Statistics on the positions of the conclusion sentences within the abstracts in the CAS-human dataset. The minus sign denotes the positions counting from the end of the abstract.}
    \label{fig:con-pos-stat}
\end{figure}

%% file: table/results-human.tex
\begin{table}[!htb]
\centering
\resizebox{\columnwidth}{!}{
\begin{tabular}{lllll}
\toprule
CAS-human & $P_k \downarrow$
  & WD $\downarrow$
  & $\text{Jaccard} \uparrow$
  & ROUGE $\uparrow$ \\
\midrule
Random-base & .4894 & .5695 & .0544 & .1765  \\
Random-plus & .2167  & .3831 & .3986  & .5490 \\
TextTilling & .2555  & .3444 & .4971 & .6153  \\
SBERT-sim & .3009 & .4199 & .4491 & .5935  \\
\midrule
$\text{GreedyCAS-base}^{10}$ & .1937 & .3089 & .5670 & .6631    \\
$\text{GreedyCAS-NN}^{12}$ & $\mathbf{.1605}$ & $\mathbf{.2543}$ & $\mathbf{.6020}$ & $\mathbf{.6668}$ \\
\bottomrule
\end{tabular}
}
\caption{Segmentation results on the CAS-human dataset. $\downarrow$ indicates that the lower the value is, the better the performance, whereas $\uparrow$ means the opposite. The superscripts of GreedyCAS approaches indicate the best empirical batch size. The best results are statistically significantly better than the closest baseline (Wilcoxon signed-ranked test).}
\label{tab:results-human}
\end{table}

%% file: table/results-auto.tex
\begin{table}[!htb]
\centering

\resizebox{\columnwidth}{!}{
\begin{tabular}{lllll}
\toprule
CAS-auto & $P_k \downarrow$ & WD $\downarrow$ & $\text{Jaccard} \uparrow$  & ROUGE $\uparrow$ \\
\midrule
Random-base & .4744 & .4890 & .1534 & .2856  \\
Random-plus & .2197 & .2452 & .3328 & .5211 \\
TextTilling & .2742  & .3131 & .4009 & .5271   \\
SBERT-sim & .1930 & $\mathbf{.2101}$ & $\mathbf{.6013}$ & $\mathbf{.7274}$  \\
\midrule
$\text{GreedyCAS-base}^{8}$ & .1656 & .2341 & .4878 & .5836  \\
$\text{GreedyCAS-NN}^{12}$ & $\mathbf{.1652}$ & .2317 & .4830 & .5717  \\
\bottomrule
\end{tabular}
}
\caption{Segmentation results on the CAS-auto dataset. The best results are statistically significantly better than the closest approach.}
\label{tab:results-auto}
\end{table}

%% file: figure/correlation.tex
\begin{figure}[!htb]
    \centering
    \includegraphics[width=\columnwidth]{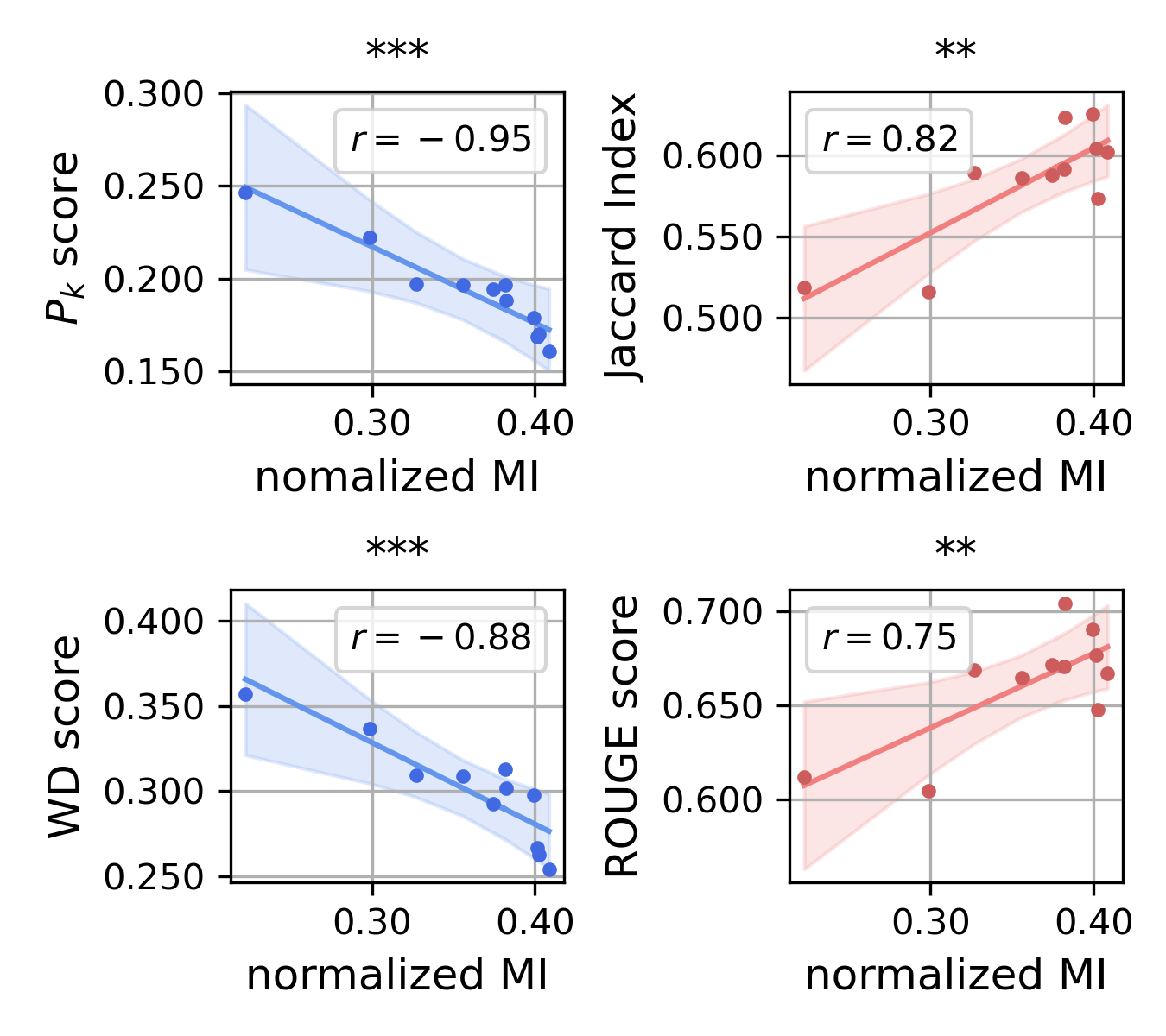}
    \caption{Correlation coefficients between NMI and other metrics. We fit linear regression models to the data points. *** indicates the significance level $p<10^{-3}$ and ** $p<10^{-2}$ (Pearson correlation test).}
    \label{fig:correlation}
\end{figure}

%% file: figure/delta-mi.tex
\begin{figure}[!htb]
    \centering
    \includegraphics[width=0.95\columnwidth]{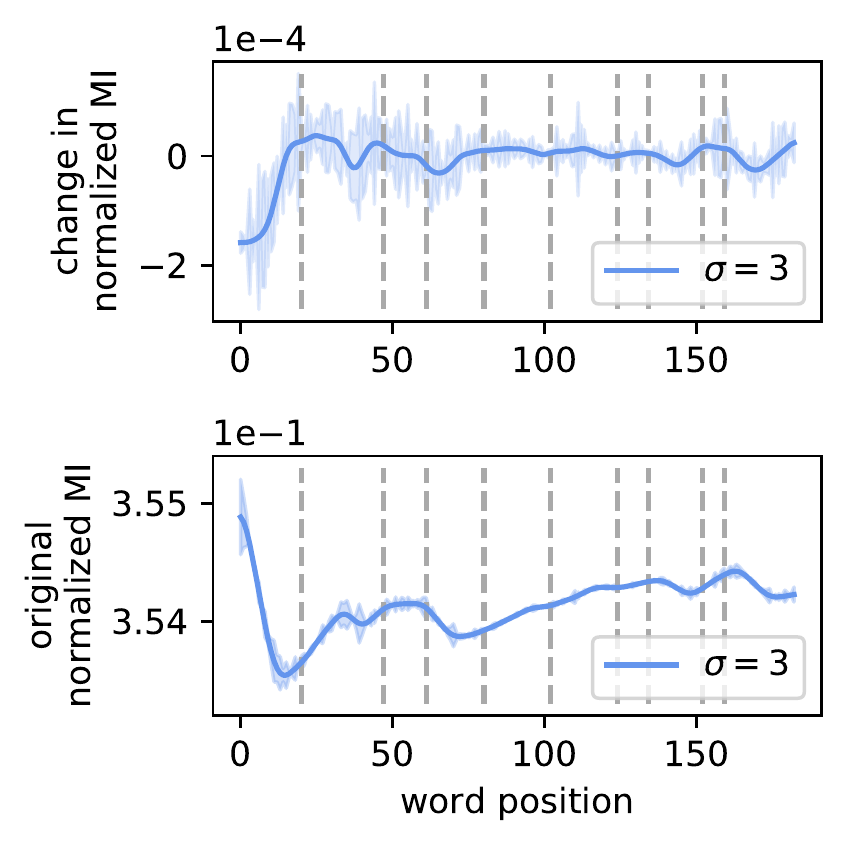}
    \caption{Change in $\text{NMI}(\mathbb{P};\mathbb{C})$ when moving one word from the premise to the conclusion segment for a fixed abstract. Dashed lines denote the end positions of sentences in the abstract. We smooth the data with a Gaussian filter with $\sigma=3$.}
    \label{fig:delta-mi}
\end{figure}

%% file: table/structured-example-A.tex
\definecolor{novemberblue}{RGB}{138, 166, 237}
\definecolor{cardinal}{RGB}{234, 158, 154}

\newcommand{\hlc}[2][yellow]{{%
    \colorlet{foo}{#1}%
    \sethlcolor{foo}\hl{#2}}%
}

\begin{table*}[!htb]
    \centering
    \begin{tabular}{p{\textwidth}}
    \toprule
    Title: Additional evidence on the efficacy of different Akirin vaccines assessed on \textit{Anopheles arabiensis} (Diptera: \textit{Culicidae}), by \citet{letinic2021additional} \\
    \midrule 
    \textit{\textbf{Background}} \hspace{0.5em} 
    \hlc[cardinal!50]{Anopheles arabiensis is an opportunistic malaria vector that rests and feeds outdoors, circumventing current indoor vector control methods.}
    \hlc[novemberblue!50]{Furthermore, this vector will readily feed on both animals and humans. Targeting this vector while feeding on animals can provide an additional intervention for the current vector control activities. Previous results have displayed the efficacy of using Subolesin/Akirin ortholog vaccines for the control of multiple ectoparasite infestations. This made Akirin a potential antigen for vaccine development against An. arabiensis.} \\
    
    \textit{\textbf{Methods}} \hspace{0.5em} 
    \hlc[novemberblue!50]{The efficacy of three antigens, namely recombinant Akirin from An. arabiensis, recombinant Akirin from Aedes albopictus, and recombinant Q38 (Akirin/Subolesin chimera) were evaluated as novel interventions for An. arabiensis vector control. Immunisation trials were conducted based on the concept that mosquitoes feeding on vaccinated balb/c mice would ingest antibodies specific to the target antigen. The antibodies would interact with the target antigen in the arthropod vector, subsequently disrupting its function.} \\
    
    \textit{\textbf{Results}} \hspace{0.5em} 
    \hlc[novemberblue!50]{All three antigens successfully reduced An. arabiensis and reproductive capacities, with a vaccine efficacy of 68-73\%.} \\ 
    
    \textit{\textbf{Conclusions}} \hspace{0.5em} 
    \hlc[cardinal!50]{These results were the first to show that hosts vaccinated with recombinant Akirin vaccines could develop a protective response against this outdoor malaria transmission vector, thus providing a step towards the development of a novel intervention for An. arabiensis vector control.} \\
    \bottomrule
    \end{tabular}
    \caption{Example abstract in CAS-auto segmented by GreedyCAS-NN, where the first sentence of the \textit{Background} category is attributed to the conclusion segment. Best view in color printing.}
    \label{tab:structured-example-A}
\end{table*}

%% file: table/structured-example-B.tex
\definecolor{novemberblue}{RGB}{138, 166, 237}
\definecolor{cardinal}{RGB}{234, 158, 154}

\begin{table*}[!htb]
    \centering
    \begin{tabular}{p{\textwidth}}
    \toprule
    Title: Interest in COVID-19 vaccine trials participation among young adults in China: Willingness, reasons for hesitancy, and demographic and psychosocial determinants, by \citet{sun2021interest} \\
    \midrule 
    \textit{\textbf{Background}} \hspace{0.5em} 
    \hlc[novemberblue!50]{With the demand for rapid COVID-19 vaccine development and evaluation, this paper aimed to describe the prevalence and correlates of willingness to participate in COVID-19 vaccine trials among university students in China.} \\
    
    \textit{\textbf{Methods}} \hspace{0.5em} 
    \hlc[novemberblue!50]{A cross-sectional survey with 1,912 Chinese university students was conducted during March and April 2020. Bivariate and multivariate analyses were performed to identify variables associated with willingness to participate.} \\
    
    \textit{\textbf{Results}} \hspace{0.5em} 
    \hlc[novemberblue!50]{The majority of participants (64.01\%) indicated willingness to participate in COVID-19 vaccine trials. Hesitancy over signing informed consent documents, concerns over time necessary for participating in a medical study, and perceived COVID-19 societal stigma were identified as deterrents, whereas lower socioeconomic status, female gender, perception of likely COVID-19 infection during the pandemic, and COVID-19 prosocial behaviors were facilitative factors.} 
    \hlc[cardinal!50]{Further, public health mistrust and hesitancy over signing informed consent documents had a significant interactive effect on vaccine trial willingness.} \\ 
    
    \textit{\textbf{Conclusions}} \hspace{0.5em} 
    \hlc[cardinal!50]{High standards of ethical and scientific practice are needed in COVID-19 vaccine research, including providing potential participants full and accurate information and ensuring participation free of coercion, socioeconomic inequality, and stigma. Attending to the needs of marginalized groups and addressing psychosocial factors including stigma and public health mistrust may also be important to COVID-19 vaccine development and future uptake.} \\
    \bottomrule
    \end{tabular}
    \caption{Example abstract in CAS-auto segmented by GreedyCAS-NN, where the last sentence of the \textit{Results} category is attributed to the conclusion segment. Best view in color printing}
    \label{tab:structured-example-B}
\end{table*}

%% file: table/word-pair-example.tex
\definecolor{novemberblue}{RGB}{138, 166, 237}
\definecolor{cardinal}{RGB}{234, 158, 154}

\begin{table*}[!htb]
    \centering
    \begin{tabular}{p{\textwidth}}
    \toprule
    Title: Hepatitis B surface antigen assembles in a post-ER, pre-Golgi compartment \cite{huovila1992hepatitis} \\
    \midrule
   \hlc[novemberblue!50]{Expression of hepatitis B surface antigen (\textbf{HBsAg}), the major envelope protein of the virus, in the absence of other viral proteins leads to its secretion as oligomers in the form of disk-like or tubular lipoprotein particles. The observation that these lipoprotein particles are heavily disulphide crosslinked is paradoxical since \textbf{HBsAg} assembly is classically believed to occur in the ER, and hence in the presence of high levels of protein disulphide isomerase (PDI) which should resolve these higher intermolecular crosslinks. Indeed, incubation of mature, highly disulphide crosslinked \textbf{HBsAg} with recombinant PDI causes the disassembly of HBsAg to dimers. We have used antibodies against resident ER proteins in double immunofluorescence studies to study the stages of the conversion of the \textbf{HBsAg} from individual protein subunits to the secreted, crosslinked, oligomer. We show that \textbf{HBsAg} is rapidly sorted to a post-ER, pre-Golgi compartment which excludes PDI and other major soluble resident ER proteins although it overlaps with the distribution of rab2, an established marker of an intermediate compartment. Kinetic studies showed that disulphide-linked \textbf{HBsAg} dimers began to form during a short (2 min) pulse, increased in concentration to peak at 60 min, and then decreased as the dimers were crosslinked to form higher oligomers. These higher oligomers are the latest identifiable intracellular form of \textbf{HBsAg} before its secretion (t 1/2 = 2 h). Brefeldin A treatment does not alter the localization of \textbf{HBsAg} in this PDI excluding compartment, however, it blocks the formation of new oligomers causing the accumulation of dimeric \textbf{HBsAg}.} 
   \hlc[cardinal!50]{Hence this oligomerization must occur in a pre-Golgi \textbf{compartment}. These data support a model in which rapid dimer formation, catalyzed by PDI, occurs in the ER, and is followed by transport of dimers to a pre-Golgi \textbf{compartment} where the absence of PDI and a different lumenal environment allow the assembly process to be completed.} \\
    \midrule
    $(w_p, w_c)$ pair that contributes the most to the overall NMI($\mathbb{P};\mathbb{C}$) score: (\textbf{HBsAg}, \textbf{compartment}) \\
    \bottomrule 
    \end{tabular}
    \caption{Example abstract in CAS-human segmented by GreedyCAS-NN. The word pair that contributes the most to $\text{NMI}(\mathbb{P};\mathbb{C})$ is in bold. Best view in color printing.}
    \label{tab:CAS-human-example}
\end{table*}